\documentclass[10pt,twocolumn,letterpaper]{article}

\usepackage[pagenumbers]{wacv}      % To produce the REVIEW version for the applications track
\definecolor{wacvblue}{rgb}{0.21,0.49,0.74}
\usepackage[pagebackref,breaklinks,colorlinks]{hyperref}

\def\wacvPaperID{3186} % *** Enter the WACV Paper ID here
\def\confName{WACV}
\def\confYear{2026}

\title{Generation of Synthetic Fingerphotos with GANs}

\author{
    Conor Miller-Lynch$^*$ \quad Sandip Purnapatra$^*$ \quad Syed Konain Abbas$^\dagger$ \quad Lambert Igene$^\dagger$ \\
    Faraz Hussain$^\dagger$ \quad Soumyabrata Dey$^\dagger$ \quad Stephanie Schuckers${^*}{^\dagger}$ \\[1.5ex]
    $^*$UNC Charlotte, Charlotte, NC, USA \\
    $^\dagger$Clarkson University, Potsdam, NY, USA \\
    {\tt\small cmillerl@charlotte.edu}
}

\begin{document}
\maketitle
\begin{abstract}
Contactless fingerprinting is an emerging approach to biometric authentication that allows users to scan their fingerprints without touching a scanner.
Due to the limited amount of contactless fingerprint data available and the security risks associated with sharing real individuals' fingerprints, it is valuable to explore methods of generating synthetic data that can be used in place of --- or in conjunction with --- real data to develop and evaluate contactless fingerprinting systems.
In this paper, we present and evaluate synthetic fingerphotos generated using StyleGAN2-ADA and StyleGAN3, existing image generation architectures.
We evaluate the realism, privacy preservation, and variety of the synthetic fingerphotos by comparing their biometric feature statistics to those of real fingerphotos, computing match scores between real and synthetic fingerphotos, and computing match scores between different synthetic fingerphotos.
This paper provides a quantitative comparison point for future evaluations of synthetic fingerphotos.
The evaluation code is made available at \url{https://github.com/cmillerlynch/fingerphoto-gan}.
\end{abstract}
    
\begin{figure}[ht]
    \centering
    \includegraphics[width=0.2\linewidth]{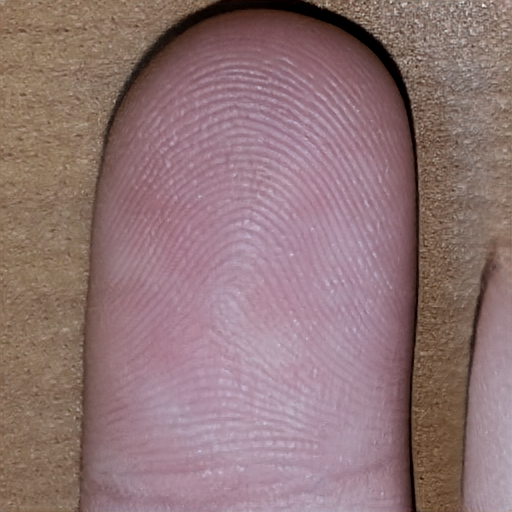}
    \includegraphics[width=0.2\linewidth]{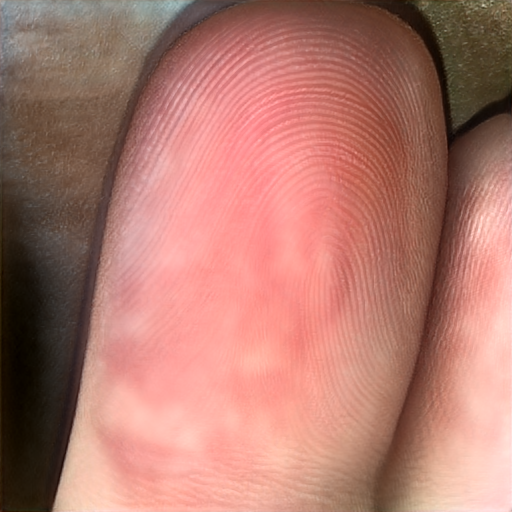}
    \includegraphics[width=0.2\linewidth]{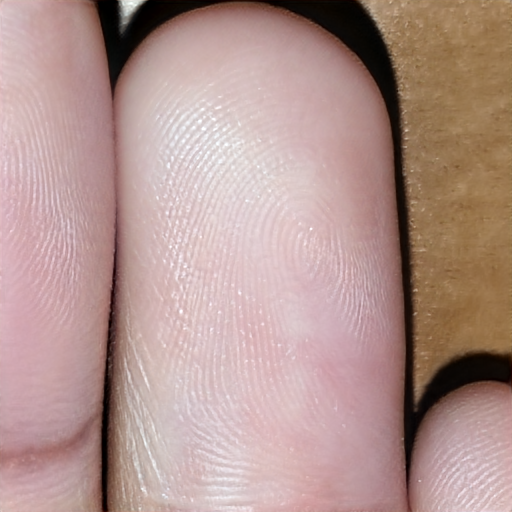}
    \includegraphics[width=0.2\linewidth]{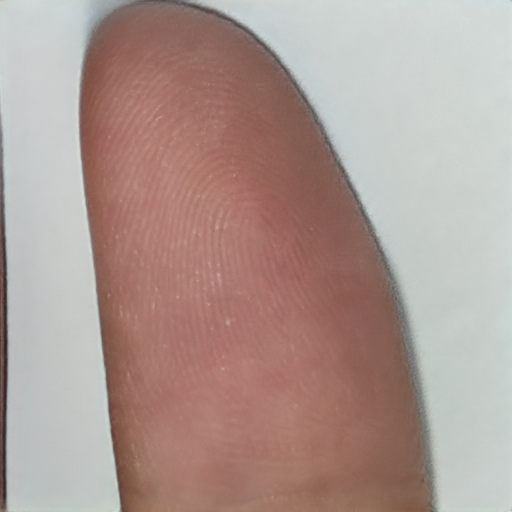}
    \includegraphics[width=0.2\linewidth]{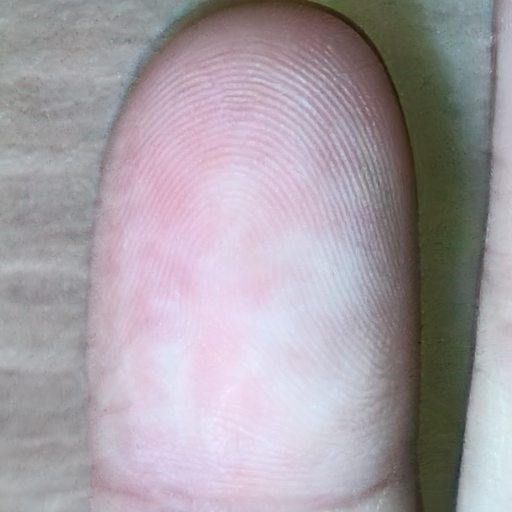}
    \includegraphics[width=0.2\linewidth]{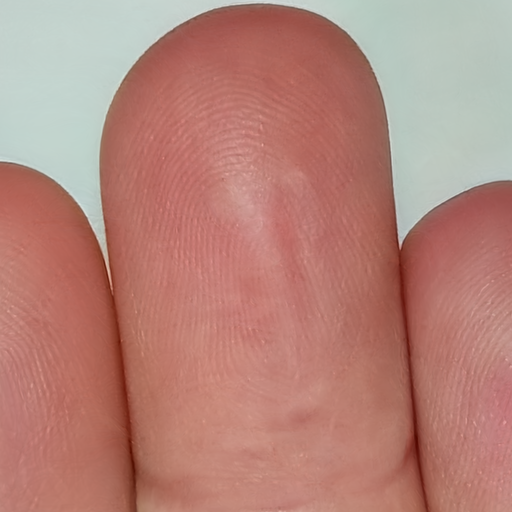}
    \includegraphics[width=0.2\linewidth]{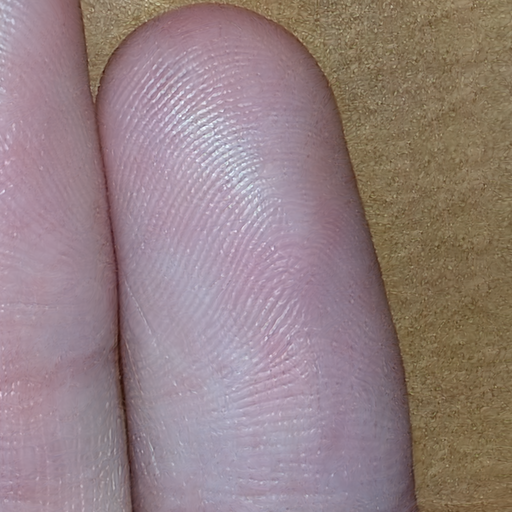}
    \includegraphics[width=0.2\linewidth]{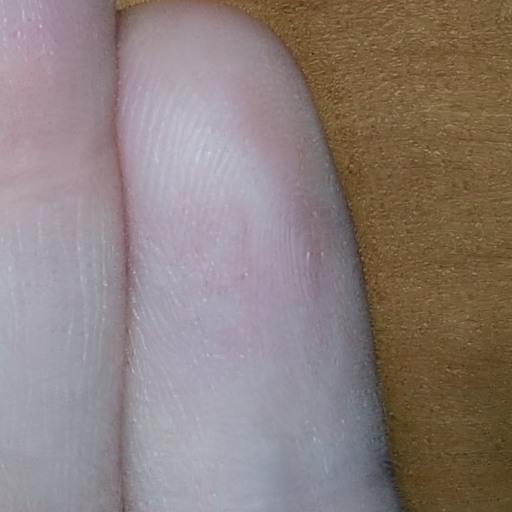}
    \caption{Synthetic fingerphotos generated using StyleGAN2-ADA (top) and StyleGAN3 (bottom). From left to right, the generated images represent index, middle, ring, and little fingers from left hands.}
    \label{synthetic_fingerphotos}
\end{figure}

\section{Introduction}

Fingerprint-based identification and authentication are widely used due to the high levels of uniqueness and permanence of human fingerprints.
However, many fingerprint identification systems require users to touch a surface to scan their fingerprints, which can be inconvenient and unsanitary.
These factors motivate the development of \textit{contactless} fingerprinting systems that match the performance of contact-based systems while being more convenient and sanitary.

To develop and compare contactless fingerprinting systems, datasets containing fingerphotos are needed.
However, fingerprint datasets cannot be widely distributed due to privacy and security risks for the individuals whose fingerprints appear in the datasets.
Furthermore, few contactless fingerprint datasets exist, and those that do often include far fewer images than their contact-based counterparts.
These factors limit the development of contactless fingerprinting systems.

Synthetic data offer a solution to both the availability and scale problems of real data, while reducing privacy and security risks because they do not represent real people.
Additionally, synthetic data can be generated on demand rather than during time-consuming collections from human subjects.

In this work, we train two end-to-end deep learning models for generation of photorealistic fingerphotos, and evaluate the generated images.
We used StyleGAN2-ADA \cite{stylegan2-ada} and StyleGAN3 \cite{stylegan3} --- existing image generation architectures --- to generate synthetic fingerphotos.

To evaluate these synthetic fingerphotos, we first develop and evaluate several image processing pipelines that can be applied to fingerphotos to allow their use with a contact-based matcher (due to the lack of publicly-available contactless matchers).
Following an approach similar to Bahmani \etal \cite{quality_uniqueness_privacy}, we compare the distributions of biometric features of the synthetic fingerphotos to those of the training images to show that the generated fingerprints have realistic biometric features.
We then evaluate the privacy preservation of the synthetic fingerphotos by comparing them to those in the training dataset, and evaluate the variety of the synthetic fingerphotos by comparing them to one another.
As part of this evaluation, we use a new visualization method that can be more informative than histograms of match scores.

% The code used to produce the results of this paper is provided on GitHub\footnote{GitHub repository URL omitted to maintain anonymity.}, allowing its use for future creation and evaluation of synthetic fingerphotos.

\section{Related Work}

Many works focus on the generation of synthetic contact-based fingerprints.
Capelli \etal \cite{synthetic_fingerprint-image_generation} proposed one of the first fingerprint generation techniques. Capelli \cite{sfinge} later proposed SFinGe, one of the most widely-used fingerprint generators.
Zhao \etal \cite{statistical_feature_models} proposed a method that uses statistical analysis to produce more realistic fingerprints.

In recent years, the rapid improvement of data-driven image generation models has led to them being incorporated into most synthetic fingerprint generators.
Cao and Jain \cite{search_at_scale} train an autoencoder, then use its decoder to initialize the generator of an I-WGAN \cite{i-wgan}.
Wyzykowski \etal \cite{level_three} generate fingerprint patterns using a modified version of Anguli \cite{anguli} (itself an implementation of SFinGe \cite{sfinge}, then add realistic texture using a CycleGAN \cite{cyclegan} model. Bahmani \etal \cite{quality_uniqueness_privacy} unconditionally generate fingerprints using a StyleGAN \cite{stylegan} model. Engelsma \etal \cite{printsgan} first create a fingerprint pattern using a BigGAN \cite{biggan} model, perform distortion and cropping, then perform rendering using another BigGAN model.
Sams \etal \cite{HQ-finGAN} use a StyleGAN2 \cite{stylegan2} model for the initial generation and a CycleGAN model to improve the quality of the output.
Grosz and Jain \cite{genprint} use two diffusion models \cite{ddpm}: one to generate a fingerprint pattern and another to create multiple impressions of that pattern. Both models can be conditioned on a prompt and the second model can be conditioned on a style image.

However, there are only a few works that focus on the generation of synthetic fingerphotos. Priesnitz \etal \cite{syncolfinger} propose a technique that simulates many aspects of fingerphotos, such as rotation, skin tone, lighting, and noise.
Dong and Kumar \cite{synthesis-of-multi-view-3d-fingerprints} propose a technique that creates 3D models of fingerprints that can then be projected to 2D images.
However, neither of these approaches generates photorealistic fingerphotos.
Grosz and Jain \cite{genprint} propose a technique that allows for the generation of synthetic fingerphotos and provide the matching performance of models trained on them. However, not all aspects of the synthetic fingerphotos are evaluated independently of the other modalities.

\section{Methods}
\label{sec:methods}

We used fingerphotos collected using smartphone cameras \cite{livedet-2023} to create our training dataset.
The original dataset contains 2065 four-finger images from 35 individuals captured using six smartphone models across two collection sessions.
Each image in the original dataset shows four fingers (thumb excluded), and corresponding XML files provide a rectangular bounding box for each fingertip.
For each fingertip in each image, we cropped a square region fully containing the bounding box of the fingertip.
This resulted in 8,260 single-finger images representing 280 distinct fingers.
We then resized each image to $512 \times 512$ pixels for compatibility with StyleGAN2-ADA and StyleGAN3.
To remove blurry images, we then calculated the variance of the Laplacian (a measure of sharpness \cite{Rosebrock_2015}) for each image and removed those with the lowest 10\% of values.
This resulted in 7,434 images in the final training dataset.
We then generated a JSON file describing which finger type (left index, left middle, etc.) each image represents, thus separating the images in the training dataset into eight classes.

We used the official implementations in PyTorch \cite{pytorch} of StyleGAN2-ADA \cite{stylegan2-ada} and StyleGAN3 \cite{stylegan3}.
These models are well-known for their impressive face generation results, and the earlier StyleGAN \cite{stylegan} and StyleGAN2 \cite{stylegan2} (sometimes in conjunction with other techniques) have been used for contact-based fingerprint generation \cite{quality_uniqueness_privacy}\cite{HQ-finGAN}.
We trained both models as class-conditional on the finger type (left index, left middle, etc.).

Both models were trained for 10,000 kimgs (a kimg corresponds to 1000 images shown to the discriminator).
Checkpoints were saved every 200 kimgs.
After 10,000 kimgs of training, we selected the checkpoint that achieved the lowest (best) FID score \cite{fid_score} for each architecture.
For our StyleGAN2-ADA model, the lowest FID score was 5.48, and for our StyleGAN3 model, the lowest FID score was 4.37.
It is important to note that this metric was designed for natural images and does not necessarily capture features of biometric relevance.
However, it is useful for selecting models for further evaluation.

Fingerprint matching and evaluation software is commonly used for evaluating synthetic contact-based fingerprints \cite{quality_uniqueness_privacy}\cite{printsgan}.
Several techniques for matching contactless fingerprints have been proposed \cite{cross-fingerprint}\cite{c2cl}\cite{ridgeformer}, but these have not been widely adopted yet.
Instead of using these, we developed an image processing pipeline to make the fingerphotos more compatible with established contact-based matching and evaluation software.

\begin{figure}
    \centering
    \includegraphics[width=\linewidth]{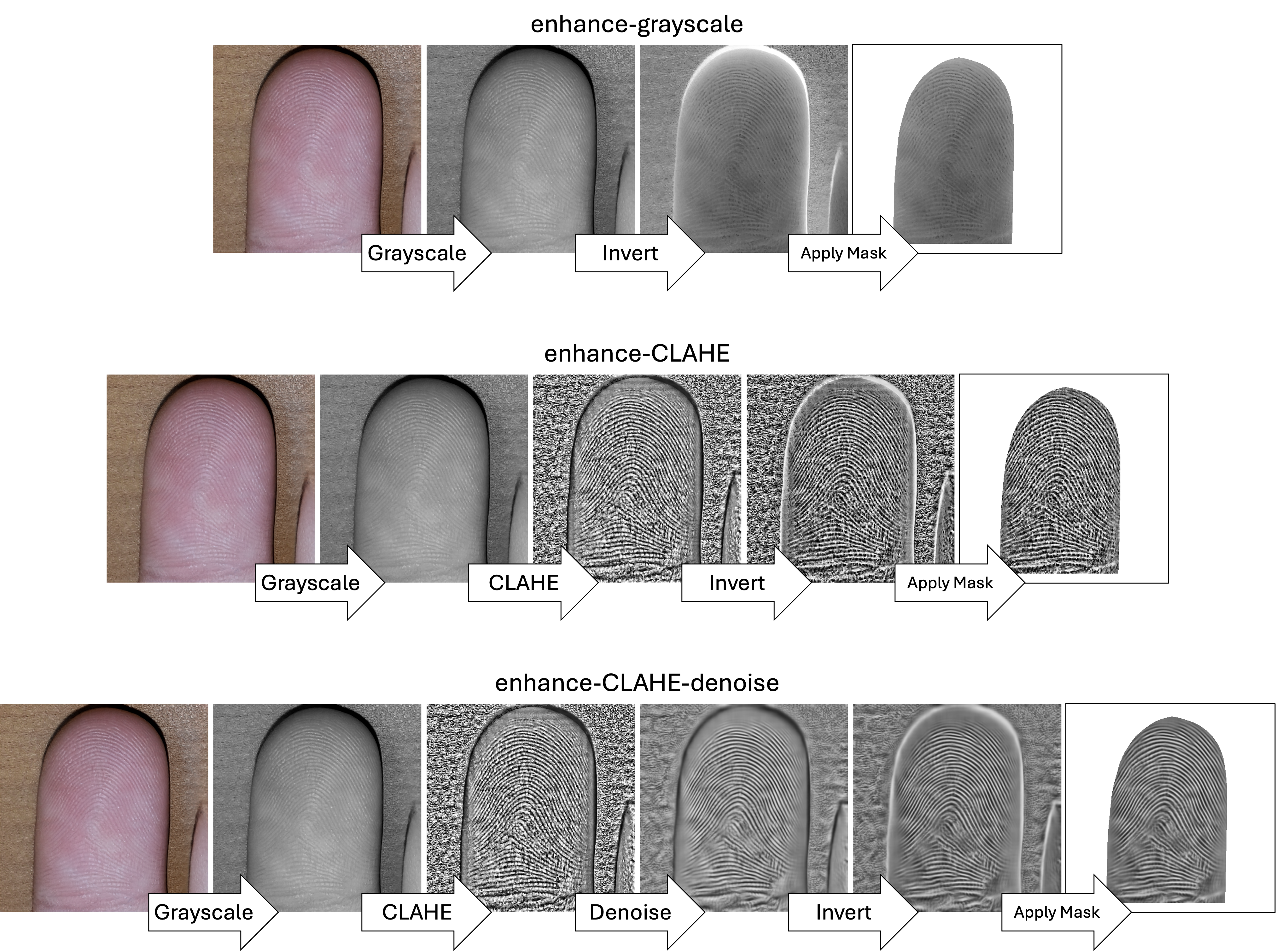}
    \caption{The image processing pipelines that were compared. enhance-CLAHE was selected to process fingerphotos for compatibility with contact-based evaluation and matching tools.}
    \label{fig:image-processing-pipelines}
\end{figure}

We evaluated three image processing pipelines using OpenCV \cite{opencv_library}:
The first pipeline (``enhance-grayscale'') converts the images to grayscale then inverts the lightness values;
the second pipeline (``enhance-CLAHE'') converts the images to grayscale, applies contrast-limited adaptive histogram equalization (CLAHE), then inverts the lightness values;
the third pipeline (``enhance-CLAHE-denoise'') converts the images to grayscale, applies CLAHE, applies non-local means denoising, then inverts the lightness values.
For each pipeline, the unprocessed images were used to generate binary masks indicating the part of each image corresponding to the finger.
These masks were then applied to the processed images, with the areas of each image outside its mask being filled with white pixels, isolating the finger from the background and adjacent fingers.
Illustrations of these pipelines are provided in \autoref{fig:image-processing-pipelines}.

\begin{figure}
    \centering
    \includegraphics[width=\linewidth]{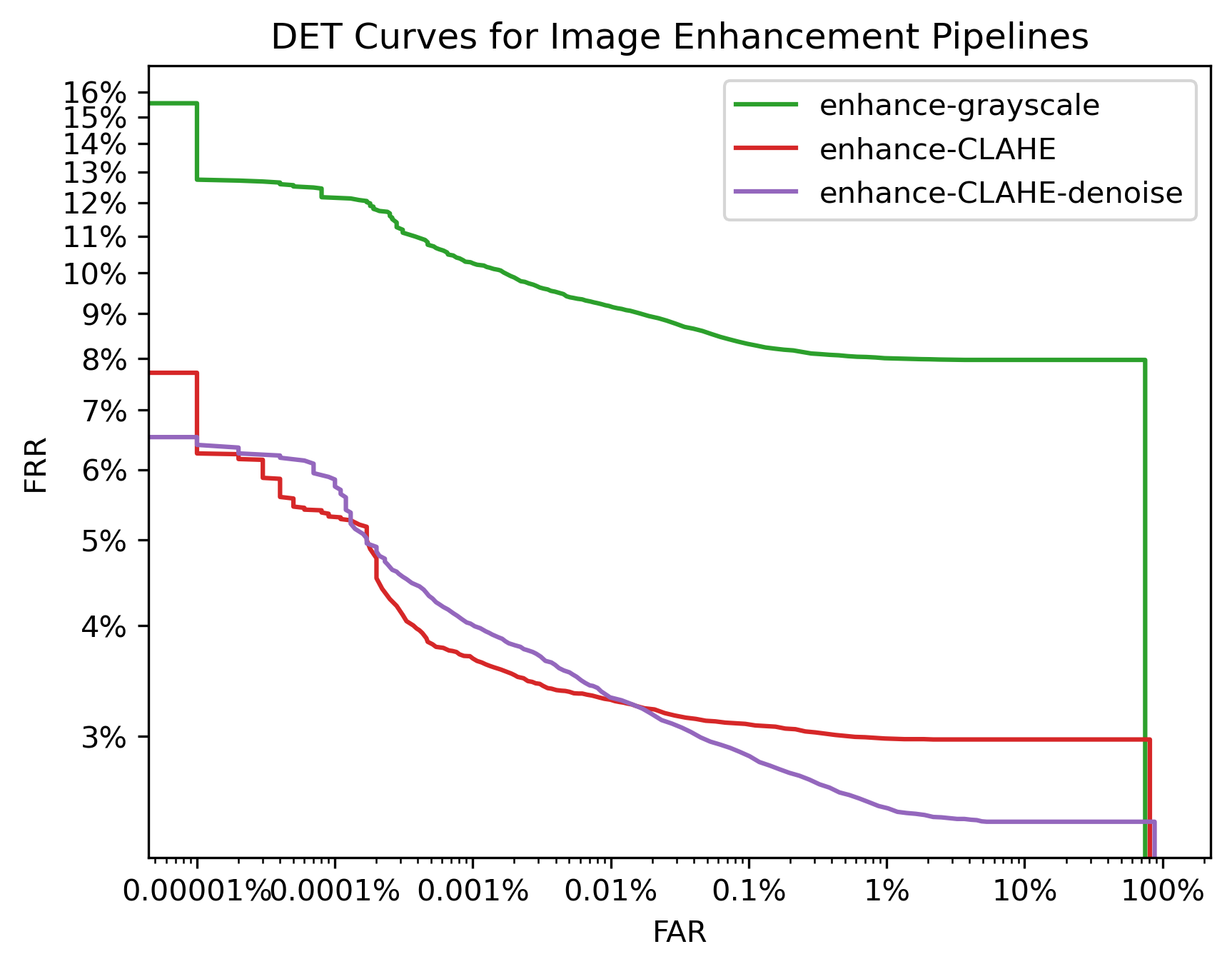}
    \caption{DET curves achieved by VeriFinger on fingerphotos processed using each image processing pipeline.}
    \label{fig:pipeline-det-curves}
\end{figure}

We compared the matching performance achieved by VeriFinger 2025.1\footnote{\url{https://neurotechnology.com/verifinger.html}} on the real fingerphotos in the training dataset to select the best pipeline.
For each pipeline, we randomly selected 1 million random pairs of processed real fingerphotos.
We excluded images of the same finger that were captured during the same collection session with the same phone, since these pairs often contain very similar images that would inflate the genuine match scores.
When a template could not be generated for one or both images in a pair, the match score for that pair was recorded as 0.
The resulting DET curves are shown in {\autoref{fig:pipeline-det-curves}}.
Note that each pipeline has a minimum FRR it can achieve due to VeriFinger's inability to extract templates for some images.
enhance-CLAHE and enhance-CLAHE-denoise offer comparable performance for practical FARs and significantly outperform enhance-grayscale. We chose to use enhance-CLAHE to evaluate the synthetic fingerphotos. This pipeline results in an FRR of 3.30\% at an FAR of 0.01\%, 3.70\% at 0.001\%, and 5.31\% at 0.0001\%.

For each of our trained models, we generated 6,250 synthetic fingerphotos of each of the eight finger types (left index, left middle, etc.), resulting in 50,000 total images per model.
We applied the enhance-CLAHE image processing pipeline to both the real and synthetic fingerphotos.

\section{Results}

To evaluate the synthetic fingerphotos, we follow an approach similar to that of \cite{quality_uniqueness_privacy} to evaluate their realism, privacy, and variety.

For each image (processed using enhance-CLAHE), we used VeriFinger 2025.1 to extract minutiae information and NIST Fingerprint Image Quality (NFIQ2) 2.3.0 \cite{nfiq2} to obtain a score indicating the fingerprint quality.
Images for which VeriFinger could not extract a template were excluded from the minutiae statistics.
Images for which NFIQ2 failed were assigned a score of 0.
The distributions of these values for real and synthetic fingerphotos can be compared and thus the biometric realism of the generated images can be assessed.
\autoref{tab:realism} shows the means and standard deviations of these distributions.
These results show that the biometric features of the synthetic fingerphotos approximate those of the real fingerphotos used for training.
The models produce images with comparable levels of biometric realism, with StyleGAN2-ADA better modeling some biometric features and StyleGAN3 better modeling others.

\begin{table*}
    \centering
    \begin{tabular}{|c|c|c|c|c|c|c|}
        \hline
        & \multicolumn{2}{|c|}{\textbf{Real}} & \multicolumn{2}{|c|}{\textbf{StyleGAN2-ADA}} & \multicolumn{2}{|c|}{\textbf{StyleGAN3}} \\
        \cline{2-7}
        & Mean & SD & Mean & SD & Mean & SD \\
        \hline
        \textbf{Mean Minutiae Quality} & 63.70 & 4.18 & 62.27 & 3.90 & 61.97 & 3.96 \\
        \hline
        \textbf{Ridge Ending Minutiae Count} & 40.38 & 10.62 & 40.94 & 10.25 & 38.67 & 9.96 \\
        \hline
        \textbf{Bifurcation Minutiae Count} & 45.71 & 11.26 & 49.63 & 11.75 & 45.06 & 10.75 \\
        \hline
        \textbf{Percentage of Bifurcation Minutiae} & 53.13\% & 6.52\% & 54.82\% & 5.79\% & 53.87\% & 6.27\% \\
        \hline
        \textbf{NFIQ2 Score \cite{nfiq2}} & 29.80 & 21.85 & 30.28 & 21.82 & 27.32 & 22.21 \\
        \hline
    \end{tabular}
    \caption{Comparison of biometric features of the real fingerphotos in the training dataset and the synthetic fingerphotos generated by the StyleGAN2-ADA and StyleGAN3 models, as measured by VeriFinger and NFIQ2. The enhance-CLAHE image processing pipeline was applied to all fingerphotos prior to feature extraction.}
    \label{tab:realism}
\end{table*}

To evaluate the privacy preservation and variety of the synthetic fingerphotos, we first computed VeriFinger match scores for 10 million randomly-selected imposter pairs of real fingerphotos from the training dataset.
This provides us with an imposter distribution, i.e., the distribution of match scores we should expect if none of the compared fingerphotos are of the same finger.
If the fingerprint patterns in the synthetic fingerphotos are sufficiently distinct from those in the real fingerphotos, we should expect to see the same distribution when real fingerphotos are matched against synthetic fingerphotos.
Similarly, if the fingerprint patterns in the synthetic fingerphotos sufficiently model the variety of fingerprint patterns in the real fingerphotos, we should expect to see the same distribution when synthetic fingerphotos are matched against other synthetic fingerphotos.
Notably, the presence of a few high match scores does not necessarily indicate a problem in either case as long as their likelihood does not exceed that of the same scores in the real imposter distribution.

\begin{figure}
    \centering
    \includegraphics[width=\linewidth]{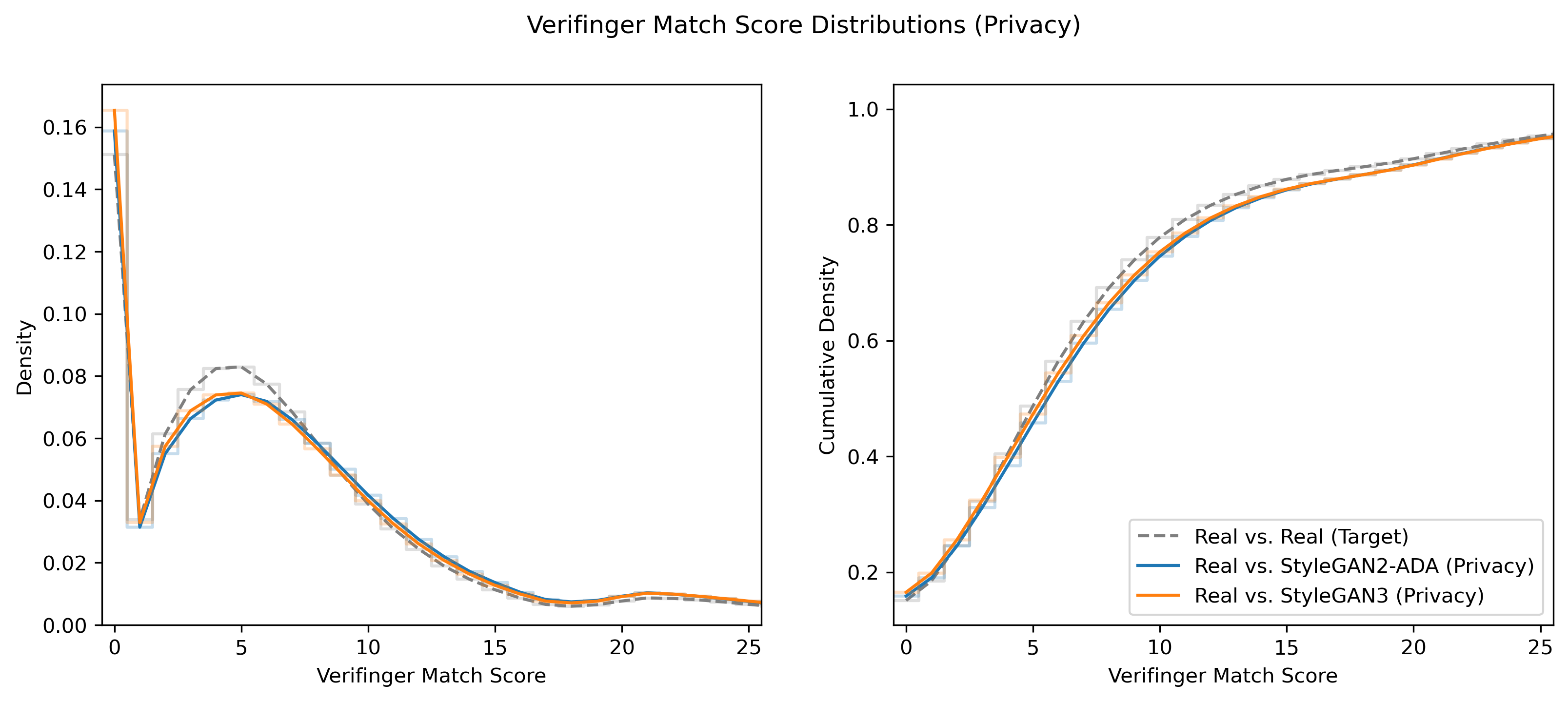}
    \caption{
    VeriFinger match score distributions for privacy evaluation.
    The more similar the real and synthetic distributions, the more distinct the synthetic fingerprint patterns are from the real ones in the training dataset.
    }
    \label{fig:privacy-preservation}
\end{figure}

\begin{figure}
    \centering
    \includegraphics[width=\linewidth]{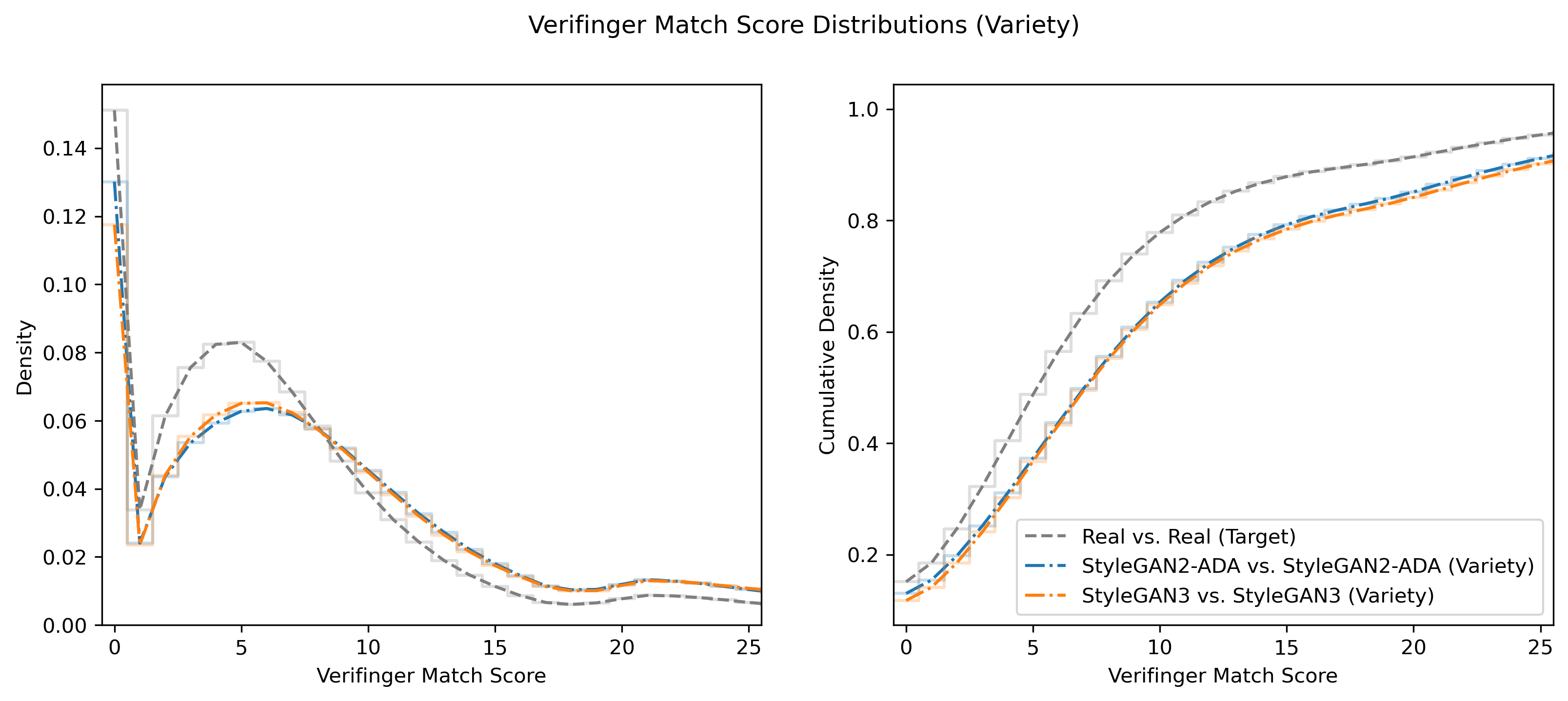}
    \caption{
    VeriFinger match score distributions for variety evaluation.
    The more similar the real and synthetic distributions, the better the synthetic fingerphotos model the variety of fingerprint patterns in the training dataset.
    }
    \label{fig:variety}
\end{figure}

For the privacy evaluation of each model, we computed VeriFinger match scores for 10 million randomly-selected pairs of images, with each pair consisting of one real fingerphoto from the training dataset and one synthetic image.
For the variety evaluation of each model, we computed VeriFinger match scores for 10 million randomly-selected pairs of images, with each pair consisting of two synthetic images generated using different seeds.
The enhance-CLAHE image processing pipeline was applied to all fingerphotos prior to matching.
When a template could not be generated for one or both images in a pair, the match score for that pair was recorded as 0.
The resulting distributions of scores are shown in \autoref{fig:privacy-preservation} and \autoref{fig:variety}.

\begin{figure}
    \centering
    \includegraphics[width=\linewidth]{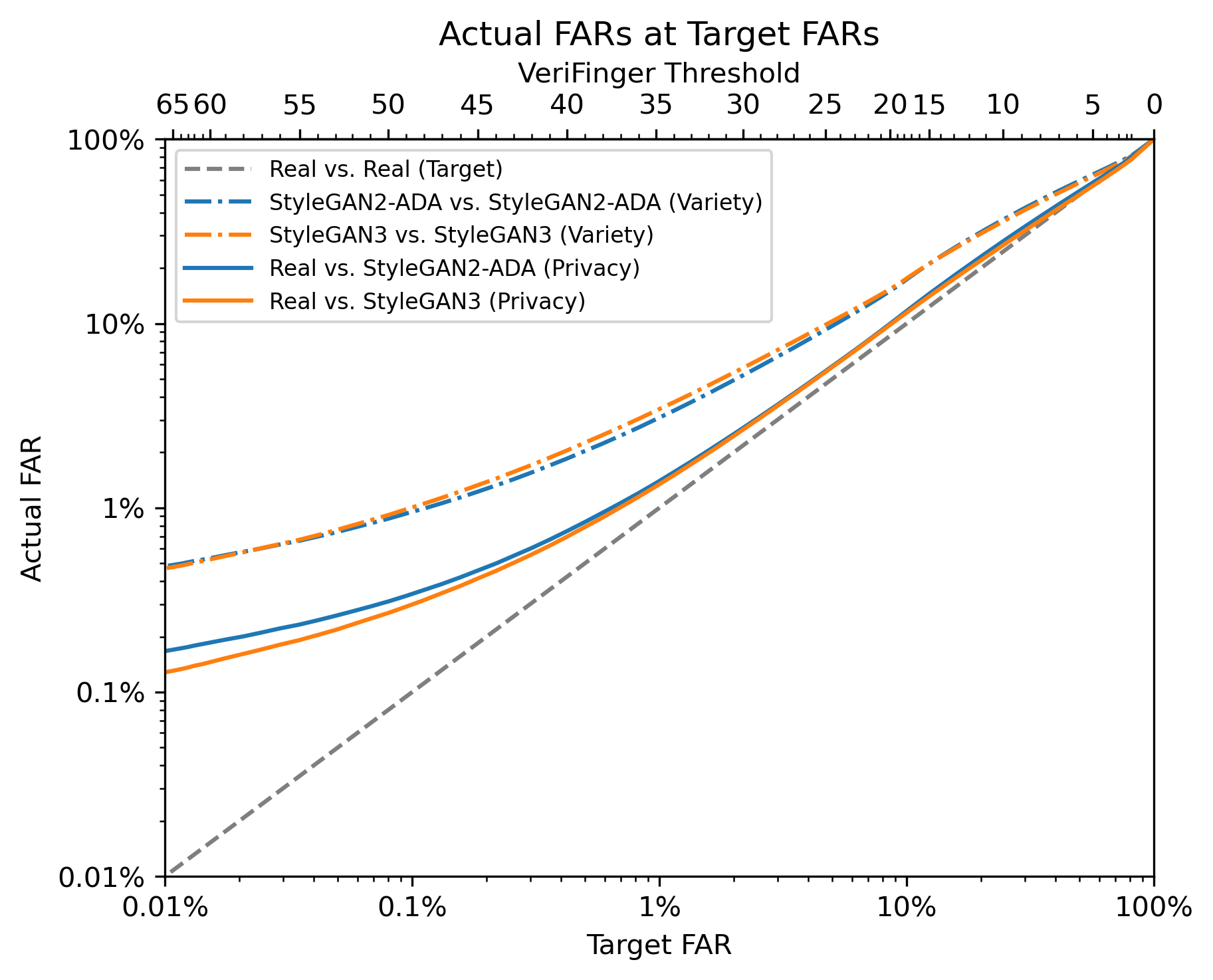}
    \caption{
    A visualiztion that compares the false accept rates for synthetic fingerphotos at a given threshold to the expected false accept rate for real fingerphotos.
    The closer each curve is to the diagonal line, the better the corresponding model achieves the corresponding goal.
    This visualization method provides more information about matching results and the tails of the distributions than histograms.
    }
    \label{fig:actual_fars_at_target_fars}
\end{figure}

While this approach of comparing match score distributions to a target imposter distribution is useful, visualizing the results as histograms can obscure valuable information, such as how these scores translate into matching results and how the tails of these distributions compare.
Therefore, we also provide an additional visualization in \autoref{fig:actual_fars_at_target_fars} that compares the false accept rates at a given threshold to the expected false accept rate for real fingerphotos.
This visualization shows that the differences between the distributions of match scores for real and synthetic fingerphotos (especially in terms of privacy) are more significant than they appear from the histograms.
This visualization is similar to evaluation methods that record the number of pairs that achieve match scores above a certain threshold \cite{quality_uniqueness_privacy}\cite{printsgan}\cite{genprint}, but expands on this idea by showing the results for a range of thresholds, giving a more complete understanding of the distributions of scores.

\section{Conclusion}

This paper provides an evaluation of synthetic fingerphotos generated using StyleGAN2-ADA and StyleGAN3.
The results of this paper can serve as a quantitative comparison point for future evaluations of synthetic fingerphotos.
The provided code and visualization methods can also be used for these evaluations.

\section{Future Work}

There are several areas for improvement that should be explored in future works.
The models cannot generate several fingerphotos of the same synthetic finger, and do not allow for the conditioning of the generation on a specific fingerprint pattern.
Furthermore, the models cannot generate fingerphotos resembling different fingers of the same subject that were captured under the same conditions.
This would require changing the finger type and pattern between images while preserving features such as skin color, lighting, and capture quality across images.
The models were trained on a relatively small dataset of fingerphotos with relatively few distinct fingers.
Increasing the dataset size and number of distinct fingers could improve the quality, privacy, and variety of the synthetic fingerphotos.
Using more recent image generation architectures may also improve the results.
This work used standard contact-based tools on processed fingerphotos.
Future works should explore using recently-developed tools specifically designed for use with fingerphotos.

\section{Acknowledgments}

This material is based upon work supported by the Center for Identification Technology Research and the National Science Foundation under Grant No. \#2413228 and \#2601332.
Results presented in this paper were obtained using the Chameleon testbed supported by the National Science Foundation.

{
    \small
    \bibliographystyle{ieeenat_fullname}
    \bibliography{main}
}

\end{document}